\pdfoutput=1

\documentclass[11pt]{article}

\usepackage[final]{acl2023}

\usepackage{times}
\usepackage{latexsym}
\usepackage{natbib}
\usepackage[T1]{fontenc}

\usepackage[utf8]{inputenc}

\usepackage{microtype}

\usepackage{inconsolata}

\usepackage{graphicx}

\usepackage{tabularx}
\usepackage{algorithm} 
\usepackage{algpseudocode} 
\usepackage{graphicx}
\usepackage{amsmath}
\usepackage{booktabs}
\usepackage{multirow}
\usepackage[export]{adjustbox}
\usepackage{makecell}
\usepackage{caption}
\usepackage{subcaption}

\title{Patentformer: A demonstration of AI-assisted automated patent drafting}

\author{Sai Krishna Reddy Mudhiganti\textsuperscript{1} \\
  \texttt{s.mudhiganti@samsung.com} \\\And
  Juanyan Wang\textsuperscript{1} \\
  \texttt{juanyan.wang@partner.samsung.com} \AND
  Ruo Yang\textsuperscript{1}  \\
  \texttt{r.yang@partner.samsung.com} \\\And
  Manali Sharma\textsuperscript{1}  \\
  \texttt{manali.s@samsung.com} \\ \AND
  \textsuperscript{1}Samsung Semiconductor, Inc. \\
  San Jose, CA \\
  \\}

\begin{document}
\maketitle
\begin{abstract}
Patent drafting presents significant challenges due to its reliance on the extensive experience and specialized expertise of patent attorneys, who must possess both legal acumen and technical understanding of an invention to craft patent applications in a formal legal writing style. This paper presents a demonstration of Patentformer, an AI-powered automated patent drafting platform designed to support patent attorneys by rapidly producing high-quality patent applications adhering to legal writing standards.
\end{abstract}

\section{Introduction}
\begin{figure*}[!h]
    \centering
    \includegraphics[scale=0.65]{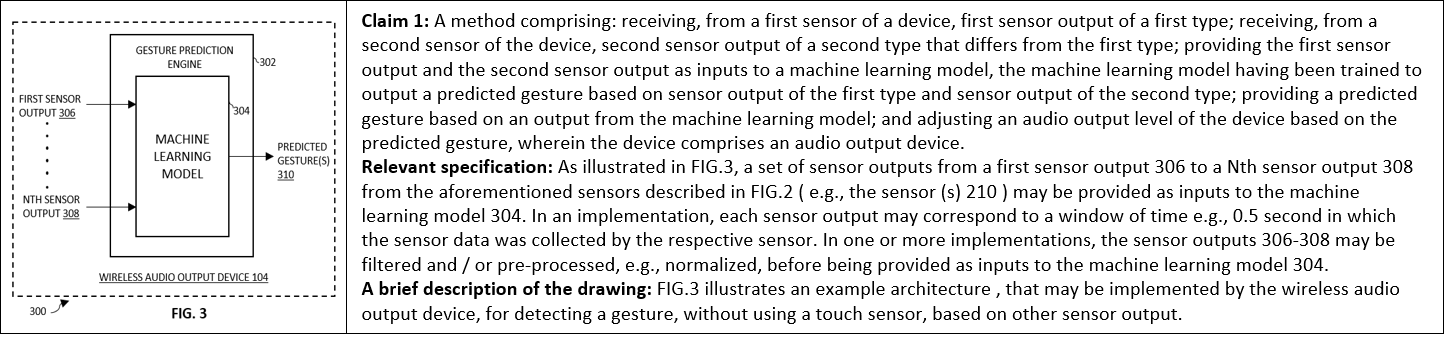}
    \caption{An example of a patent drawing (left), claim, specification, and a brief description of the drawing (right).}
    \label{fig:claims_to_specification}
\end{figure*}

Patents are legal documents that adhere to a precise writing style, where specific terminology has distinct meanings. For example, an “embodiment” of an invention refers to its concrete implementation or physical manifestation. A patent document typically comprises of several organized sections, including the title, abstract, field of the invention, background, summary, independent claims, dependent claims, drawings, a brief description of the drawings, and a detailed description of the invention, commonly referred to as the specification. Traditionally, patents are drafted by patent attorneys who possess extensive expertise in both legal principles and the patent system. The cost of drafting a moderately complex patent is, on average, over \$10K \cite{patentCost}. Patent attorneys typically review invention disclosure documents and conduct interviews with the inventor(s) to gain a comprehensive understanding of the technical details of the invention. They subsequently draft the claims, drawings, and specification. The claims define the legal boundaries of the invention, necessitating expertise in legal and technical interpretation. Drawings must adhere to the patent office’s requirements and must label each element referenced in the specification with a unique identifier. However, the most substantial portion of a patent document is the specification, which requires significant effort to draft as it provides a detailed description of the invention in accordance with the claims and drawings. Figure \ref{fig:claims_to_specification} presents an example illustrating the relationship between a patent claim, its corresponding drawing, and the supporting specification. In this work, we assume that the patent attorneys can provide their drafted claims along with supplementary drawings as input to our system for automated processing.

Transformer-based Large Language Models (LLMs) such as BERT \cite{devlin2018bert}, T5 \cite{raffel2020exploring}, Gemini \cite{team2023gemini}, GPT-3 \cite{brown2020gpt3}, and GPT-4 \cite{achiam2023gpt4}, have demonstrated remarkable capabilities in natural language generation. However, generating high-quality patent specifications remains a significant challenge for these models. Patent documents differ fundamentally from general text due to their complex legal and technical nature. Each claim must be explicitly supported by the specification, which, in turn, must provide a detailed and structured description of the invention, often referencing associated drawings. Unlike general web text, patents contain domain-specific language, intricate relationships among the claims and drawing descriptions, and extensive technical details, making it difficult for the LLMs to generate accurate and legally compliant specifications.

Additionally, patent specifications are typically lengthy, spanning multiple pages, which poses a challenge for most LLMs that are constrained by fixed token limits (e.g., 512, 1024, 2048, or 4096 tokens). These limitations hinder the ability of models to generate coherent and contextually rich patent text. Another major issue is that most pretrained LLMs are not trained on patent data, resulting in poor adaptation to the precise writing style and legal requirements of patent documents. As a result, standard LLMs struggle to produce text that aligns with the stylistic and structural conventions required for patent specification drafting.

In this paper, we provide demonstration of Patentformer \cite{wang2024patentformer}, a novel system designed to generate patent specifications, and address the challenges outlined earlier. Patentformer operates by taking a patent claim and any associated drawing text as input. The system first preprocesses and enhances the input to improve its readability and structure, facilitating a better comprehension by the LLMs. The enhanced text is then passed to a fine-tuned LLM, which has been specifically trained on publicly available patent data to learn the stylistic and structural conventions of patent writing. This enables the model to generate high-quality patent specifications that align with legal and technical standards. Finally, we deploy Patentformer as an interactive patent drafting assistant, providing the users with an intuitive interface to streamline the patent writing process. The main contributions of this paper include the following:
\begin{itemize}
    \itemsep0em
    \item We developed and deployed Patentformer, a platform designed to assist in patent specification drafting, at \url{https://patentformer.com}. The system accepts a patent claim and its corresponding drawing text as input from the user and generates high-quality patent specifications, ensuring coherence with legal and technical requirements.\footnote{The username and password for accessing the system are `acldemo2025'. We provide one example of input claims and drawings with the supplementary materials to test the Patentformer App. The video demonstration is available at \url{https://www.youtube.com/watch?v=2_JaJoUUVGM}.}
    \item We developed a specialized training data construction methodology that transforms plain text into an enriched representation, significantly enhancing the quality of outputs generated by Patentformer. The dataset contains 1,006,494 samples which were processed from the G06F CPC category of patents available from the USPTO\footnote{\url{https://www.uspto.gov/web/patents/classification/cpc/html/cpc-G06F.html}}. We publicly release this dataset at \url{https://huggingface.co/datasets/rypatentformer/patentformer_dataset2/tree/main}.
    \item We conducted a user study to quantitatively evaluate the effectiveness of Patentformer in generating patent specifications. The results demonstrate its capability to produce high-quality and legally coherent specifications.
\end{itemize}
\section{Related Work}
Previous research on patent text generation has primarily focused on generating specific sections of a patent rather than the entire specification. For instance, \citet{lee2020patent} fine-tuned GPT-2 to generate patent claims, while \citet{lee2020patentper} extended this approach by incorporating a BERT-based module to personalize claim generation. \citet{lee2019measuring} introduced a span-based method and a quantitative framework for evaluating patent claim generation, and \citet{jiang2024patentclaims} proposed a method for generating patent claims from detailed descriptions.

Several studies have explored structured text-to-text transformations for patent text generation. \citet{lee2020controlling} leveraged structural metadata within patent documents to guide text generation, where specific keywords in the input dictated different generation tasks. \citet{lee2020measuringsem} investigated semantic search-based methods for controlling patent text generation. \citet{lee2023evaluating} trained the GPT-J model from scratch using patent corpora for an autocompletion task and introduced the Autocomplete Effectiveness (AE) ratio as a new evaluation metric. Building on this, \citet{jieh2022effectiveness} further enhanced GPT-J-6B by pre-training it bidirectionally on patent text data. Additionally, \citet{christofidellis2022pgt} developed the Patent Generative Transformer (PGT), a GPT-2-based model designed for part-of-patent generation tasks.

Another research direction has focused on summarizing patent text to generate concise outputs, such as titles \cite{souza2021comparative}, abstracts \cite{guoliang2023generating, zhu2023automatic}, prior art summaries \cite{lee2020prior}, and captions for patent figures \cite{aubakirova2023patfig}. The closest related work is the research by \citet{jiang2024patentclaims} that generated claims from the specification, which is the reversed direction of our objective.

While these approaches contribute to patent text processing, they do not address the challenge of generating complete and coherent patent specifications. To the best of our knowledge, Patentformer is the first platform that generates high-quality patent specifications from claims and drawing text.

\section{Background on Patentformer}


The objective of Patentformer algorithm \cite{wang2024patentformer} is to generate detailed rich text based on the provided textual information. Formally, let $\mathcal{P}$ represent a patent document consisting of:
\begin{itemize}
    \itemsep0em
    \item A sequence of $l$ claims, denoted as $\mathcal{C} = \{c_1, c_2, \dots, c_l\}$.
    \item A sequence of $m$ specification paragraphs, denoted as $\mathcal{S} = \{s_1, s_2, \dots, s_m\}$.
    \item A set of $t$ drawing images, denoted as $\mathcal{I} = \{i_1, i_2, \dots, i_t\}$.
    \item  A set of $t$ brief descriptions of the drawings, denoted as $\mathcal{B} = \{b_1, b_2, \dots, b_t\}$, where each $b_r$ corresponds to an image $i_r \in \mathcal{I}$.
\end{itemize}

For each drawing image $i_z \in \mathcal{I}$, let $n_z$ represent a set of $k$ pairs of component names and their corresponding component numbers appearing in the drawing. Formally, we define: $n_z$ = \{<$i_{z_1}^{name}$, $i_{z_1}^{num}$>,<$i_{z_2}^{name}$, $i_{z_2}^{num}$>, ..., <$i_{z_k}^{name}$, $i_{z_k}^{num}$>\}, where $i_{z_j}^{name}$ denotes the name of the $j^{th}$ component, and $i_{z_j}^{num}$ represents its corresponding number in the image $i_z$. The complete set of component name-number pairs across all images is denoted as $\mathcal{N} = \{n_1, n_2, \dots, n_t\}$.

The task is to generate the output specification $\mathcal{S}$ using the claims $\mathcal{C}$, the drawing descriptions $\mathcal{B}$, and the component name-number pairs $\mathcal{N}$ as inputs. The generated specification must: i) incorporate and support all the features present in the claims $\mathcal{C}$, ii) accurately describe the drawings using the drawing descriptions $\mathcal{B}$, and iii) correctly reference the components in the drawings by utilizing the component name-number pairs in $\mathcal{N}$. This process is formally expressed as $\mathcal{T} \rightarrow \mathcal{S}$, where $\mathcal{T}$ represents the combined input information $\{\mathcal{C}, \mathcal{B}, \mathcal{N}\}$ used to generate the output specification $\mathcal{S}$. Instead of using the plain text for the training tuples, e.g., $\mathcal{T}$ = ($\mathcal{C}$, $\mathcal{B}$, $\mathcal{N}$), Patentformer designs a rich version of $\mathcal{T}$ as  $\mathcal{T'}$ = ($\mathcal{C'}$, $\mathcal{B'}$, $\mathcal{N'}$), and generates the processed $\mathcal{S'}$. That is, the Patentformer targets the task of $\mathcal{T'} \rightarrow \mathcal{S'}$, instead of $\mathcal{T} \rightarrow \mathcal{S}$.

\section{Patentformer System}
\begin{figure*}[!h]
    \centering
    \includegraphics[scale=.15]{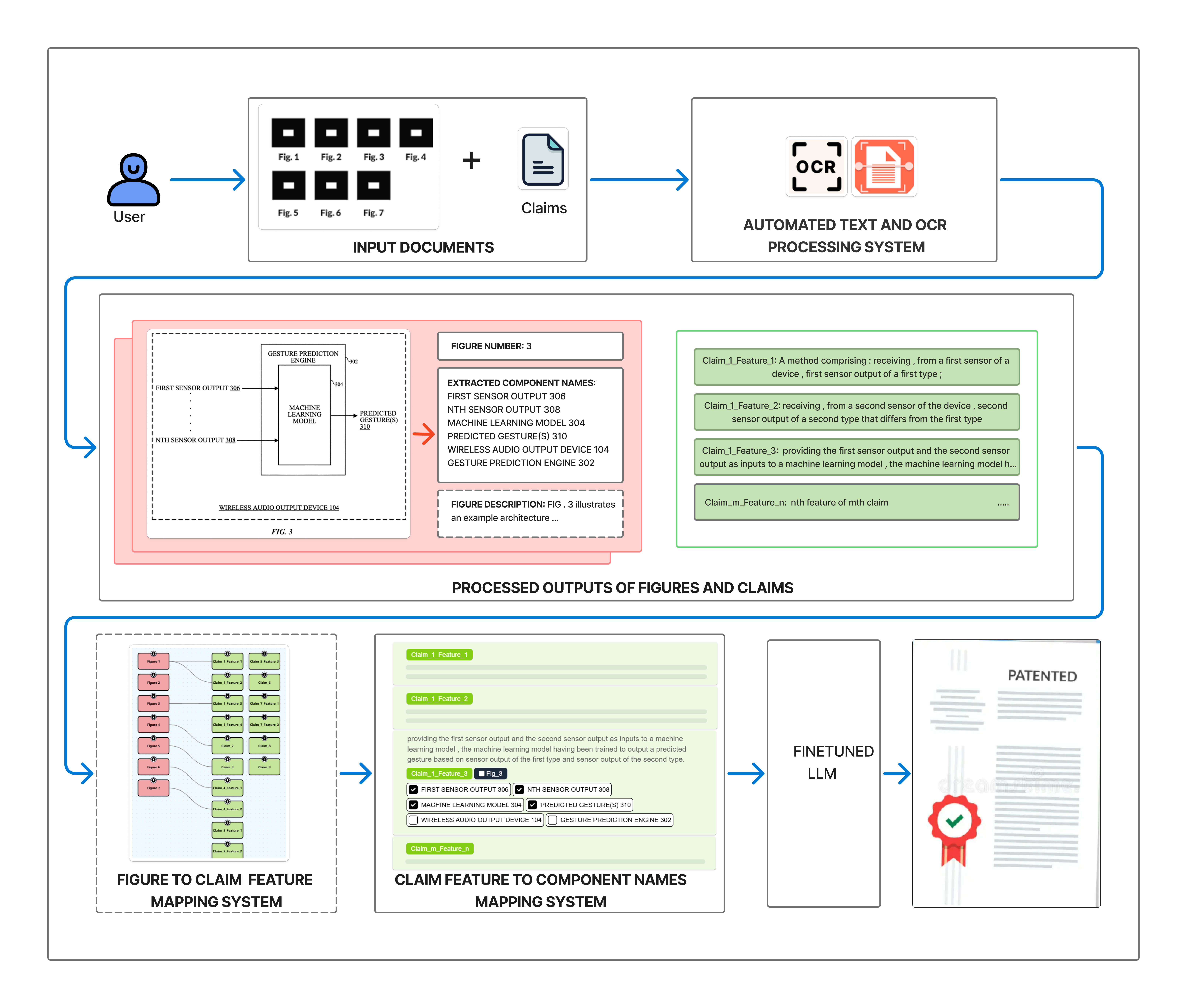}
    \caption{Workflow of the Patentformer system. The user provides the claims $\mathcal{C}$, and the images $\mathcal{I}$ with descriptions $\mathcal{B}$ to the Patentformer, then the Patentformer processes the inputs automatically to an enhanced text version, e.g., $\mathcal{C'}$, $\mathcal{N'}$, and $\mathcal{S'}$, as described in \cite{wang2024patentformer}. Furthermore, a user interface is provided to the user to map the correct relationship between claims with components and descriptions. Thereafter, the system automatically prepares the linked mapping model input $\mathcal{T'}$ and sends it to the fine-tuned large language model. Finally, the model outputs the generated specifications $\mathcal{S'}$ and is automatically cleaned to a clean version $\mathcal{S}$ before presenting to the user.}
    \label{fig:workflow}
    
\end{figure*}

We developed the Patentformer system, which utilizes the Patentformer algorithm, and deployed a user-interactive version capable of automatically generating and outputting patent-style text based on user inputs. To generate the final output specifications, the Patentformer system follows these three steps: 1. Input the claims and extract the drawing text from Visio drawings. 2. Map the figure and its components to the corresponding claims. 3. The underlying LLM generates specifications based on the input claims and drawing text pairs. Figure \ref{fig:workflow} illustrates the workflow of the Patentformer system.

\begin{figure*}[!h]
    \centering
     \begin{subfigure}[b]{.525\textwidth}
         \centering
         \includegraphics[width=\textwidth]{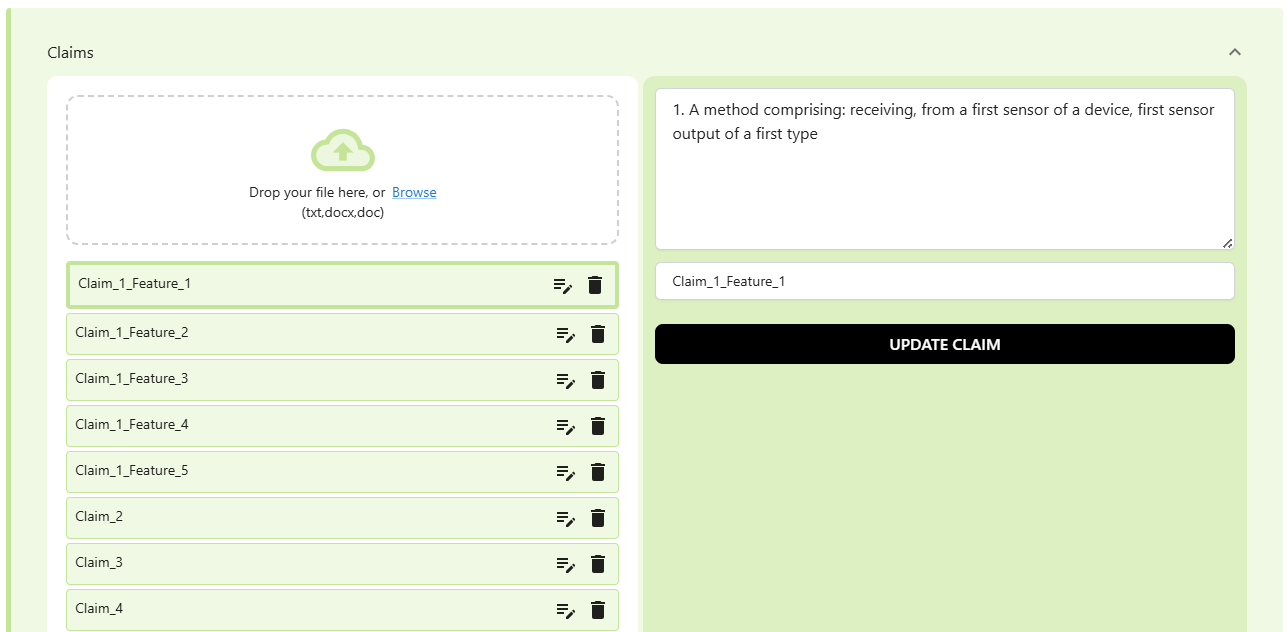}
         \caption{GUI to input and preprocess claims}
         \label{fig:claim_input}
     \end{subfigure}
     \hfill
     \begin{subfigure}[b]{.44\textwidth}
         \centering
         \includegraphics[width=\textwidth]{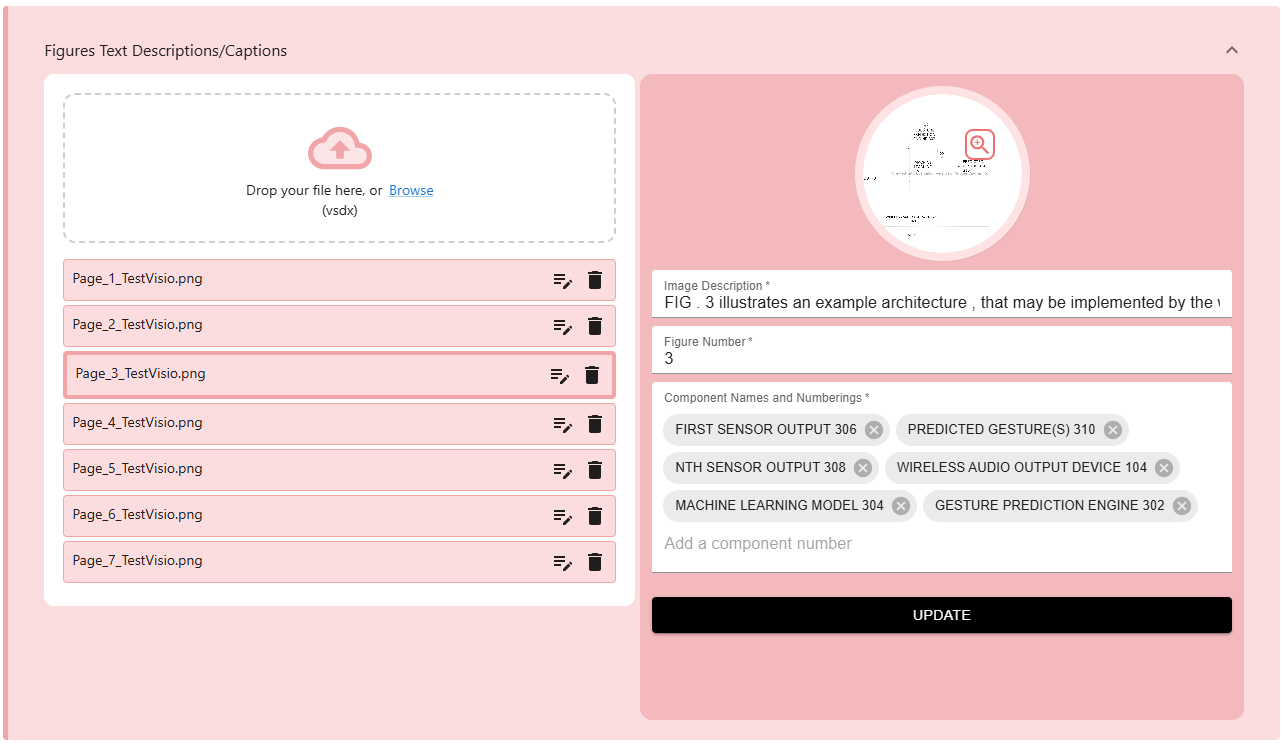}
         \caption{GUI to input and preprocess drawings (figures)}
         \label{fig:Drawing_input}
     \end{subfigure}
     \begin{subfigure}[b]{.49\textwidth}
         \includegraphics[width=\textwidth]{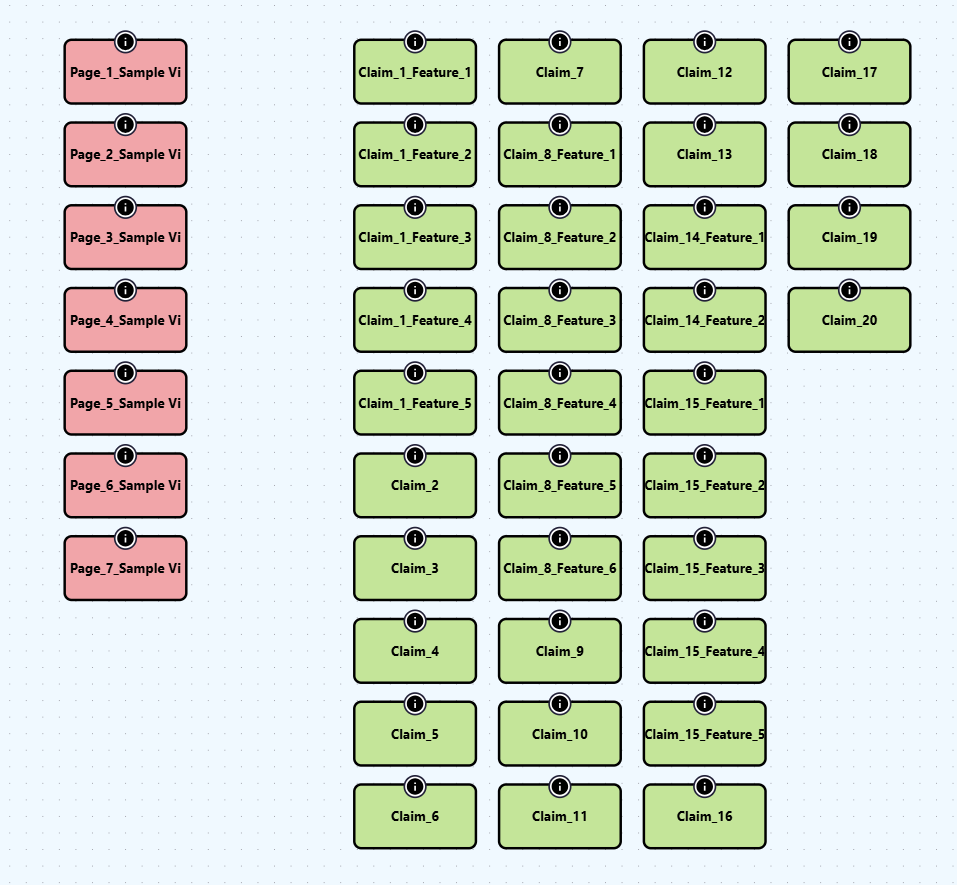}
         \caption{Unmapped claim features and drawings in GUI}
         \label{fig:unmap}
     \end{subfigure}
     \hfill
     \begin{subfigure}[b]{.48\textwidth}
         \includegraphics[width=\textwidth]{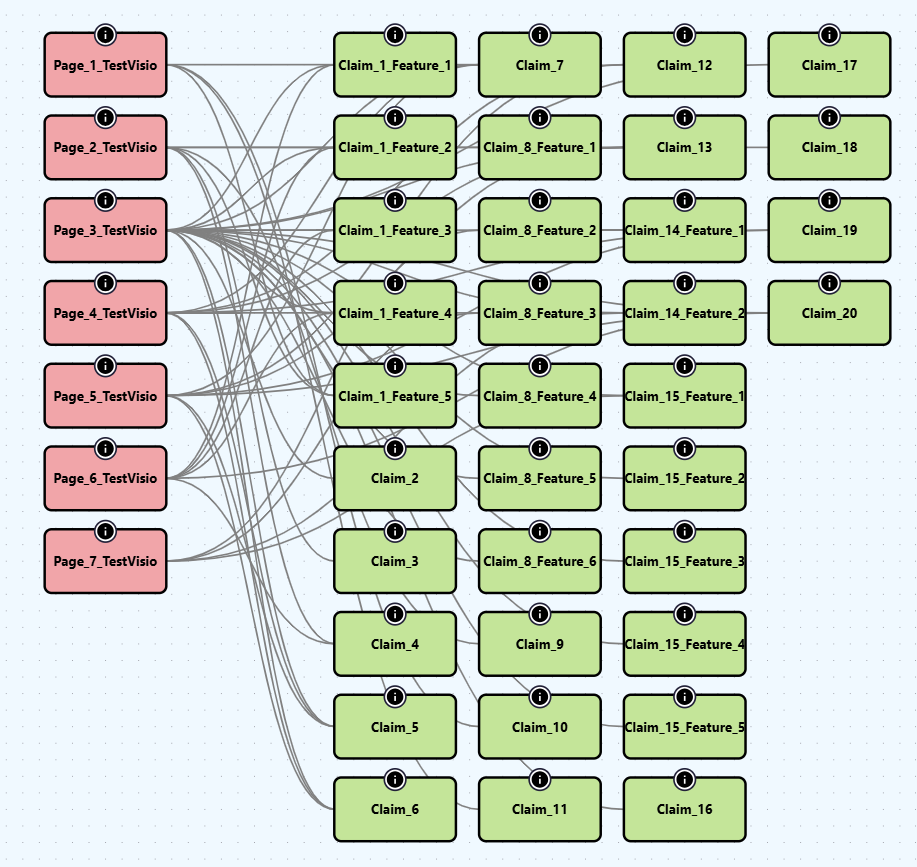}
         \caption{Mapped claim feature and drawing pairs in GUI}
         \label{fig:mapped}
     \end{subfigure}
     \begin{subfigure}[b]{\textwidth}
         \includegraphics[width=\textwidth]{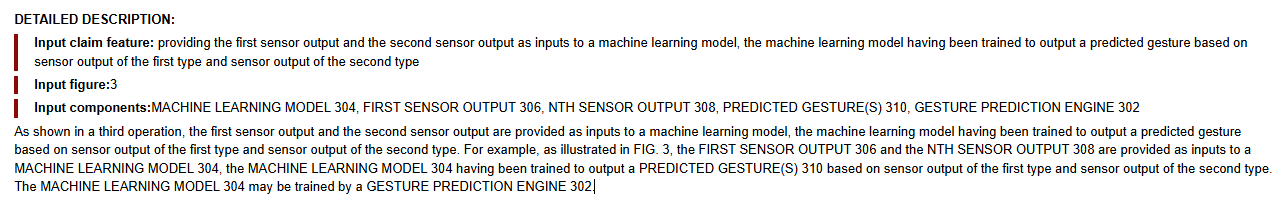}
         \caption{An example output from the Patentformer}
         \label{fig:output}
     \end{subfigure}
     
     \caption{User interface and output of the Patentformer system from input to generated specification.}

\end{figure*}

\subsection{Inputs}

The process begins with the user uploading the claim text, $\mathcal{C}$, and the corresponding drawing figures, $\mathcal{I}$, to the Patentformer platform. As illustrated in Figures \ref{fig:claim_input} and \ref{fig:Drawing_input}, the interface requires the user to provide these inputs and the Patentformer processes the text and figures. This preprocessing step includes generating structured claim features from the claim text and automatically identifying key components and their respective numbers, denoted as $\mathcal{N'}$, within the drawings. Additionally, the user has the option to modify the processed text and manually add an image description, $\mathcal{B}$, through the interactive interface. Once these inputs are finalized, Patentformer further refines and enhances them, resulting in transformed representations such as the enhanced claim text $\mathcal{C'}$, extracted components $\mathcal{N'}$, and the enriched image description $\mathcal{B'}$.

\subsection{Mapping}
Once Patentformer has processed the text, the platform presents a Mapping Interface, allowing the users to define relationships among various claims and drawing features, including components and descriptions. As illustrated in Figure \ref{fig:unmap}, the interface initially displays unlinked claims and drawing features. Users can then manually establish connections between these elements by specifying relationships through the user interface, as shown in Figure \ref{fig:mapped}. For example, a user may indicate that the drawing feature labeled “Page\_1\_Text\_Visio” corresponds to Claim Feature 1. This mapping process ensures that the generated patent specification accurately aligns claims with their respective visual components. On the other hand, we implemented a simple strategy to automatically match the components to claims feature based on cosine similarity, and BLEU-1 and BLEU-2 scores; we used a threshold of 0.1 to select up to top five matching components with each claim feature.\footnote{This strategy resulted in  precision@5 of 0.565 and  precision@3 of 0.6 based on approximately 6,000 samples.}

\subsection{Output}

Finally, each mapped claim, along with its associated components, is processed into a structured text input tuple $\mathcal{T'}$ = ($\mathcal{C'}$, $\mathcal{N'}$, $\mathcal{B'})$, which is then passed to the underlying fine-tuned LLM. The LLM generates an enriched version of the patent specification, $\mathcal{S'}$, for each claim. Before presenting the output to the user, a final post-processing step refines it to conform to standard patent writing conventions. Figure \ref{fig:output} provides an example of the generated patent specification produced by Patentformer.

\begin{figure}[ht]
    \centering
    \includegraphics[scale=.47]{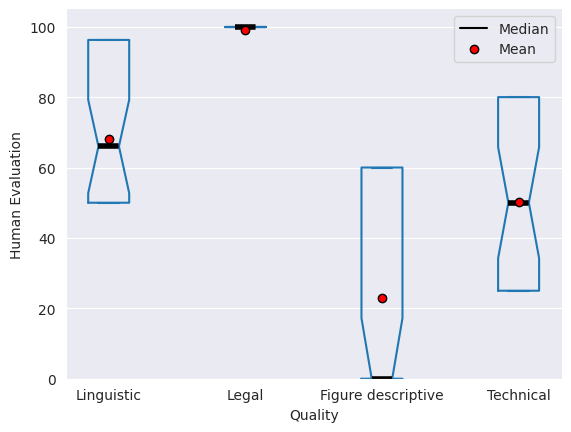}
    \caption{Human evaluation score across four qualities: linguistic, legal, figure descriptive, and technical quality. The user agrees with the legal (the most important aspect of the patent) and the linguistic quality of the drafting from the Patentformer and expects the potential from the multimodal version of the Patentformer to improve the figure descriptive and technical qualities.}
    \label{fig:user_study}
\end{figure}

\section{Experiments}
In this section, we present the experiments for evaluating Patentformer \cite{wang2024patentformer} from practical perspective.


\noindent \textbf{Dataset and model.} We constructed Patent-2015-2024-G06F, the first dataset that contains patent specifications, claims, and drawings, using USPTO patents from 2015–2024. It includes 1,006,494 training tuples. The text was truncated to 512 tokens with T5-11B \cite{raffel2020exploring} model. We further fine-tuned the model on the dataset from its pretrained version. We used the post-processing strategy defined in \cite{wang2024patentformer}.

\noindent \textbf{User study.} Patentformer \cite{wang2024patentformer} fully evaluated the effectiveness of the fine-tuned LLM, compared to the state-of-the-art baseline general LLM, for the patent generation task. To further evaluate Patentformer system in practice, we performed a user study where we asked the user, an expert patent writer with many published patents, to score the system based on their interactive experience with the platform. In particular, the user assessed the system's quality along four orthogonal dimensions: linguistic, legal, figure descriptive, and technical quality measures. We asked the user to assign a score, ranging from 0 to 100, to each quality measure, taking into account their knowledge and expertise. Figure \ref{fig:user_study} shows the scores for the four quality measures over thirty generations. As the most important aspect of drafting the patent, the legal quality is nearly full score, which also confirms that the Patentformer learns well from legal (patent) documents. Also, even if we encrypt the inputs with special tokens during the fine-tuning, e.g., <fig num>, the user is satisfied with the linguistics, which indicates that the underlying LLM still inherits the ability to generate well-written documents from the original LLM. On the other hand, the model lacks the ability to understand the given image. This behavior is expected due to two reasons: i) the current model takes image description as the input text instead of the image directly, and ii) the model is a text-to-text model, which does not have the multimodality to understand the given image. This suggests the possibility of enhancing the patent generation process by incorporating a vision-language model, rather than relying solely on the text model, which is a topic for future research. Overall, the user benefits from the Patentformer when drafting the patent due to its linguistic and legal quality, and the user expects improvements from the multimodality version of the Patentformer.

\noindent \textbf{Running time.}  To generate the specification for the user's input, Patentformer supports running the generation on GPU, and it automatically switches to CPU if no GPU is available. Table \ref{tab:runningtime} reports the average generation time, with both CPU and GPU settings, for generating one specification. We also report the average running time using GPUs for five actual use cases that were generated by different users. It is worth noting that different use cases lead to different actual running times, for example, a user may input different number of claim features and its feature mappings, where the model will generate different number of specifications. 

\begin{table}[h]
\centering
\resizebox{.48\textwidth}{!}{%
\begin{tabular}{|c|c|c|c|}
\hline
Settings & CPU only & GPU (A100) & Actual Use Case (on GPU)\\
\hline
Time (s) & $3152.4 \pm 160.00$ & $92 \pm 5.79$ & $807 \pm 449.25$\\
\hline
\end{tabular}%
}
\caption{Running times of the current Patentformer system based on CPU and GPUs on different cases.}
\label{tab:runningtime}
\end{table}
\vspace{-5mm}

\section{Conclusions and Future Work}

In this work, we introduced Patentformer, a platform designed to facilitate the drafting of patent specifications by leveraging Large Language Models (LLMs). The system processes patent claims and associated drawing descriptions, and generates detailed and legally compliant specifications. To enhance the quality of the generated text, we developed a specialized training data construction methodology that enriches plain text representations, enabling the LLM to produce more accurate and structured outputs. Through extensive quantitative evaluations, we demonstrated the effectiveness of Patentformer in generating high-quality patent specifications that align with legal and technical requirements. The results confirm the system’s potential as a valuable tool for automating and improving the patent drafting process.


The system currently relies on user-provided figure descriptions, as the underlying LLM operates solely on textual inputs and lacks direct image comprehension capabilities. To address this limitation, our ongoing research focuses on developing multimodal models, such as Large Vision-Language Models (LVLMs), that integrate both textual and visual information. These models would enable automatic interpretation of patent figures, reducing the need for manual input and further improving specification generation. Additionally, the creation of patent figures remains an expensive and labor-intensive task, as these figures must convey high-level concepts while maintaining sufficient details to support the patent document. A potential future direction involves leveraging image generative models, such as Stable Diffusion \cite{rombach2022high}, to automate the generation of patent figures based on user input, further streamlining the patent drafting process.


\newpage
\section*{Ethics Statement}
We constructed the \emph{Patent-2015-2024-G06F dataset} from publicly available patent data provided by the USPTO. The user study reviews about quality are subjective views of the patent experts. Patents are legal documents, and the USPTO\footnote{https://www.federalregister.gov/documents/2024/04/11/2024-07629/guidance-on-use-of-artificial-intelligence-based-tools-in-practice-before-the-united-states-patent} recommends that practitioners take extra care to verify the technical accuracy of the documents and compliance with 35 U.S.C. 112 when using AI drafting tools \cite{holman2024usptoAI}.

\newpage
\bibliographystyle{acl_natbib}
\bibliography{main}

\begin{thebibliography}{23}
\expandafter\ifx\csname natexlab\endcsname\relax\def\natexlab#1{#1}\fi

\bibitem[{Achiam et~al.(2023)Achiam, Adler, Agarwal, Ahmad, Akkaya, Aleman, Almeida, Altenschmidt, Altman, Anadkat et~al.}]{achiam2023gpt4}
Josh Achiam, Steven Adler, Sandhini Agarwal, Lama Ahmad, Ilge Akkaya, Florencia~Leoni Aleman, Diogo Almeida, Janko Altenschmidt, Sam Altman, Shyamal Anadkat, et~al. 2023.
\newblock Gpt-4 technical report.
\newblock \emph{arXiv preprint arXiv:2303.08774}.

\bibitem[{Aubakirova et~al.(2023)Aubakirova, Gerdes, and Liu}]{aubakirova2023patfig}
Dana Aubakirova, Kim Gerdes, and Lufei Liu. 2023.
\newblock Patfig: Generating short and long captions for patent figures.
\newblock In \emph{Proceedings of the IEEE/CVF International Conference on Computer Vision}, pages 2843--2849.

\bibitem[{Brown et~al.(2020)Brown, Mann, Ryder, Subbiah, Kaplan, Dhariwal, Neelakantan, Shyam, Sastry, Askell et~al.}]{brown2020gpt3}
Tom Brown, Benjamin Mann, Nick Ryder, Melanie Subbiah, Jared~D Kaplan, Prafulla Dhariwal, Arvind Neelakantan, Pranav Shyam, Girish Sastry, Amanda Askell, et~al. 2020.
\newblock Language models are few-shot learners.
\newblock \emph{Advances in neural information processing systems}, 33:1877--1901.

\bibitem[{Christofidellis et~al.(2022)Christofidellis, Torres, Dave, Roveri, Schmidt, Swaminathan, Vandierendonck, Zubarev, and Manica}]{christofidellis2022pgt}
Dimitrios Christofidellis, Antonio~Berrios Torres, Ashish Dave, Manuel Roveri, Kristin Schmidt, Sarath Swaminathan, Hans Vandierendonck, Dmitry Zubarev, and Matteo Manica. 2022.
\newblock Pgt: a prompt based generative transformer for the patent domain.
\newblock In \emph{ICML 2022 Workshop on Knowledge Retrieval and Language Models}.

\bibitem[{Devlin et~al.(2019)Devlin, Chang, Lee, and Toutanova}]{devlin2018bert}
Jacob Devlin, Ming-Wei Chang, Kenton Lee, and Kristina Toutanova. 2019.
\newblock \href {https://api.semanticscholar.org/CorpusID:52967399} {Bert: Pre-training of deep bidirectional transformers for language understanding}.
\newblock In \emph{North American Chapter of the Association for Computational Linguistics}.

\bibitem[{Guoliang et~al.(2023)Guoliang, Shu, Yunfeng, Chunjiang, and Liang}]{guoliang2023generating}
Shi Guoliang, Zhou Shu, Wang Yunfeng, Shi Chunjiang, and Liu Liang. 2023.
\newblock Generating patent text abstracts based on improved multi-head attention mechanism.
\newblock \emph{Data Analysis and Knowledge Discovery}, 7(6):61--72.

\bibitem[{Holman(2024)}]{holman2024usptoAI}
Christopher~M Holman. 2024.
\newblock The us patent and trademark office’s response to recent developments in artificial intelligence.
\newblock \emph{Biotechnology Law Report}.

\bibitem[{Jiang et~al.(2024)Jiang, Zhang, Scherz, and Goetz}]{jiang2024patentclaims}
Lekang Jiang, Caiqi Zhang, Pascal~A Scherz, and Stephan Goetz. 2024.
\newblock Can large language models generate high-quality patent claims?
\newblock \emph{arXiv preprint arXiv:2406.19465}.

\bibitem[{Jieh-Sheng(2022)}]{jieh2022effectiveness}
LEE Jieh-Sheng. 2022.
\newblock The effectiveness of bidirectional generative patent language models.
\newblock In \emph{Legal Knowledge and Information Systems: JURIX 2022: The Thirty-fifth Annual Conference, Saarbr{\"u}cken, Germany, 14-16 December 2022}, volume 362, page 194. IOS Press.

\bibitem[{Lee(2020{\natexlab{a}})}]{lee2020controlling}
Jieh-Sheng Lee. 2020{\natexlab{a}}.
\newblock Controlling patent text generation by structural metadata.
\newblock In \emph{Proceedings of the 29th ACM International Conference on Information \& Knowledge Management}, pages 3241--3244.

\bibitem[{Lee(2020{\natexlab{b}})}]{lee2020measuringsem}
Jieh-Sheng Lee. 2020{\natexlab{b}}.
\newblock Measuring and controlling text generation by semantic search.
\newblock In \emph{Companion Proceedings of the Web Conference 2020}, pages 269--273.

\bibitem[{Lee(2020{\natexlab{c}})}]{lee2020patentper}
Jieh-Sheng Lee. 2020{\natexlab{c}}.
\newblock Patent transformer: A framework for personalized patent claim generation.
\newblock In \emph{CEUR Workshop Proceedings}, volume 2598. CEUR-WS.

\bibitem[{Lee(2023)}]{lee2023evaluating}
Jieh-Sheng Lee. 2023.
\newblock Evaluating generative patent language models.
\newblock \emph{World Patent Information}, 72:102173.

\bibitem[{Lee and Hsiang(2020{\natexlab{a}})}]{lee2020patent}
Jieh-Sheng Lee and Jieh Hsiang. 2020{\natexlab{a}}.
\newblock Patent claim generation by fine-tuning openai gpt-2.
\newblock \emph{World Patent Information}, 62:101983.

\bibitem[{Lee and Hsiang(2020{\natexlab{b}})}]{lee2019measuring}
Jieh-Sheng Lee and Jieh Hsiang. 2020{\natexlab{b}}.
\newblock Patenttransformer-1.5: Measuring patent claim generation by span relevancy.
\newblock In \emph{New Frontiers in Artificial Intelligence}, pages 20--33, Cham. Springer International Publishing.

\bibitem[{Lee and Hsiang(2020{\natexlab{c}})}]{lee2020prior}
Jieh-Sheng Lee and Jieh Hsiang. 2020{\natexlab{c}}.
\newblock Prior art search and reranking for generated patent text.
\newblock \emph{arXiv preprint arXiv:2009.09132}.

\bibitem[{Quinn(2015)}]{patentCost}
Gene Quinn. 2015.
\newblock \href {https://ipwatchdog.com/2015/04/04/the-cost-of-obtaining-a-patent-in-the-us/id=56485/} {The cost of obtaining a patent in the us}.
\newblock Accessed: 2024-7-18.

\bibitem[{Raffel et~al.(2020)Raffel, Shazeer, Roberts, Lee, Narang, Matena, Zhou, Li, and Liu}]{raffel2020exploring}
Colin Raffel, Noam Shazeer, Adam Roberts, Katherine Lee, Sharan Narang, Michael Matena, Yanqi Zhou, Wei Li, and Peter~J Liu. 2020.
\newblock Exploring the limits of transfer learning with a unified text-to-text transformer.
\newblock \emph{The Journal of Machine Learning Research}, 21(1):5485--5551.

\bibitem[{Rombach et~al.(2022)Rombach, Blattmann, Lorenz, Esser, and Ommer}]{rombach2022high}
Robin Rombach, Andreas Blattmann, Dominik Lorenz, Patrick Esser, and Bj{\"o}rn Ommer. 2022.
\newblock High-resolution image synthesis with latent diffusion models.
\newblock In \emph{Proceedings of the IEEE/CVF conference on computer vision and pattern recognition}, pages 10684--10695.

\bibitem[{Souza et~al.(2021)Souza, Meireles, and Almeida}]{souza2021comparative}
Cinthia~M Souza, Magali~RG Meireles, and Paulo~EM Almeida. 2021.
\newblock A comparative study of abstractive and extractive summarization techniques to label subgroups on patent dataset.
\newblock \emph{Scientometrics}, 126(1):135--156.

\bibitem[{Team et~al.(2023)Team, Anil, Borgeaud, Wu, Alayrac, Yu, Soricut, Schalkwyk, Dai, Hauth et~al.}]{team2023gemini}
Gemini Team, Rohan Anil, Sebastian Borgeaud, Yonghui Wu, Jean-Baptiste Alayrac, Jiahui Yu, Radu Soricut, Johan Schalkwyk, Andrew~M Dai, Anja Hauth, et~al. 2023.
\newblock Gemini: a family of highly capable multimodal models.
\newblock \emph{arXiv preprint arXiv:2312.11805}.

\bibitem[{Wang et~al.(2024)Wang, Mudhiganti, and Sharma}]{wang2024patentformer}
Juanyan Wang, Sai Krishna~Reddy Mudhiganti, and Manali Sharma. 2024.
\newblock Patentformer: A novel method to automate the generation of patent applications.
\newblock In \emph{Proceedings of the 2024 Conference on Empirical Methods in Natural Language Processing: Industry Track}, pages 1361--1380.

\bibitem[{Zhu et~al.(2023)Zhu, Zheng, and Feng}]{zhu2023automatic}
Changsheng Zhu, Xin Zheng, and Wenfang Feng. 2023.
\newblock An automatic generation method of patent specification abstract based on" extraction-abstraction" model.
\newblock In \emph{2023 IEEE 3rd International Conference on Power, Electronics and Computer Applications (ICPECA)}, pages 196--200. IEEE.

\end{thebibliography}
\end{document}